\documentclass{article}

\usepackage[preprint]{neurips_2025}

\usepackage[utf8]{inputenc} 
\usepackage[T1]{fontenc}    
\usepackage{hyperref}       
\usepackage{url}            
\usepackage{booktabs}       
\usepackage{amsfonts}       
\usepackage{nicefrac}       
\usepackage{microtype}      
\usepackage{xcolor}         
\usepackage{makecell}
\usepackage{mathrsfs}
\usepackage{amsmath}
\usepackage{svg}
\usepackage{graphicx}
\usepackage{listings}
\usepackage{float}
\usepackage{caption}
\usepackage{multirow}
\usepackage{comment}

\newcommand{\warn}[1]{\textcolor{red}{#1}}
\title{T-Rex: Task-Adaptive Spatial Representation Extraction for Robotic Manipulation with Vision-Language Models}

\author{
Yiteng Chen\textsuperscript{1,*}, Wenbo Li\textsuperscript{1,*}, Shiyi Wang\textsuperscript{2,*},\\
Huiping Zhuang\textsuperscript{3,\dag}, Qingyao Wu\textsuperscript{1,\dag}\thanks{
  Corresponding authors: Huiping Zhuang (\href{mailto:hpzhuang@scut.edu.cn}{hpzhuang@scut.edu.cn}) and Qingyao Wu (\href{mailto:qyw@scut.edu.cn}{qyw@scut.edu.cn})
} \\
\textsuperscript{1}School of Software Engineering, South China University of Technology \\
\textsuperscript{2}School of Future Technology, South China University of Technology \\
\textsuperscript{3}Shien-Ming Wu School of Intelligent Engineering, South China University of Technology
}

\begin{document}

\maketitle

\begin{abstract}
Building a general robotic manipulation system capable of performing a wide variety of tasks in real-world settings is a challenging task. Vision-Language Models (VLMs) have demonstrated remarkable potential in robotic manipulation tasks, primarily due to the extensive world knowledge they gain from large-scale datasets. In this process, Spatial Representations (such as points representing object positions or vectors representing object orientations)  act as a bridge between VLMs and real-world scene, effectively grounding the reasoning abilities of VLMs and applying them to specific task scenarios. However, existing VLM-based robotic approaches often adopt a fixed spatial representation extraction scheme for various tasks, resulting in insufficient representational capability or excessive extraction time. In this work, we introduce T-Rex, a Task-Adaptive Framework for Spatial Representation Extraction, which dynamically selects the most appropriate spatial representation extraction scheme for each entity based on specific task requirements. Our key insight is that task complexity determines the types and granularity of spatial representations, and Stronger representational capabilities are typically associated with Higher overall system operation costs. Specifically, we design Chain of Grounding (CoG)  to guide VLMs in progressively grounding instructions, incrementally inferring all spatial constraints codes at each stage, as well as the necessary spatial representations and their optimal extraction scheme. We develop an extensible Spatial Representation Extraction Toolkit that dynamically invokes Large Vision Models according to the CoG inference results, and calls fine-grained extraction methods only when deemed necessary by VLMs. Subsequently, low level action sequence generator generates the robot action sequence based on the constraints and the tracked spatial representations. Through comprehensive experiments in real-world robotic environments, we show that our approach delivers significant advantages in spatial understanding,  efficiency, and stability without additional training.
\end{abstract}
\begin{figure}[ht]
  \centering
  \includegraphics[width=\linewidth]{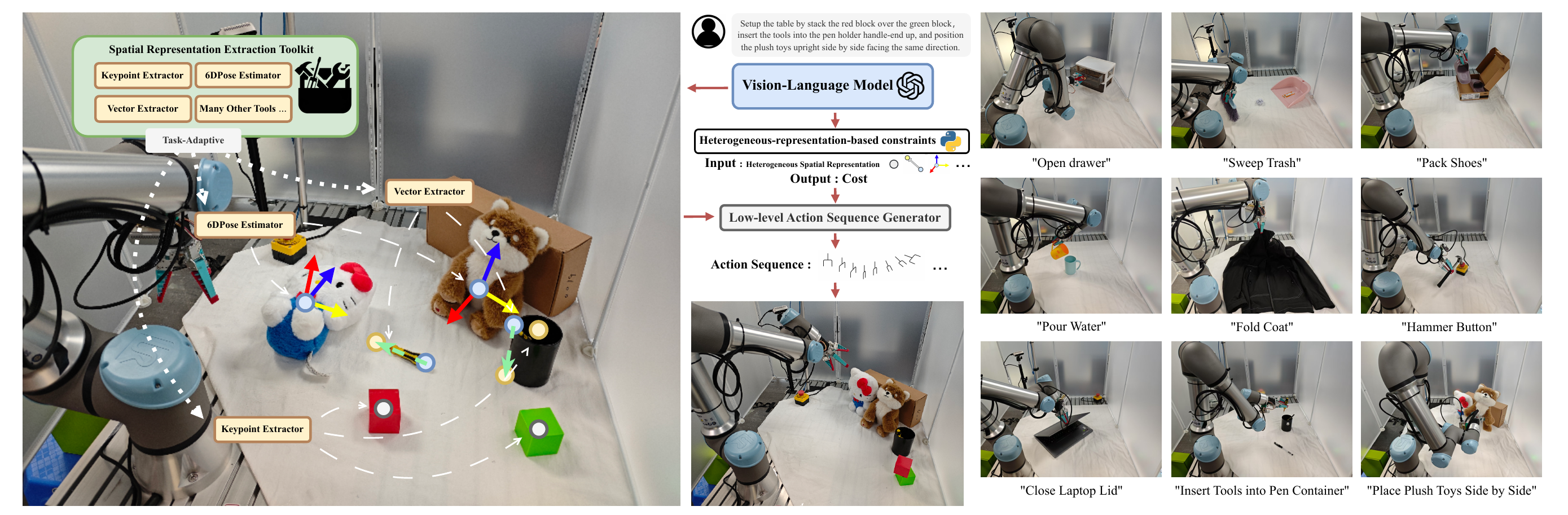}
  \caption{T-Rex adaptively invokes the optimal extractor from the Spatial Representation Extraction Toolkit to obtain spatial representations for each task-relevant object. In table organization task it extracts point for blocks, vector for tools and the pen holder, and 6D pose for plush toys, while defining positional constraints on blocks, orientation constraints on pen and pen holder, and pose constraints on plush toys.}
  \label{fig:diagram}
\end{figure}

\section{Introduction}
\label{intro}

Developing a general robotic manipulation system capable of performing tasks in complex, variable, and unstructured real-world environments has long been a challenging task. Vision-Language Models (VLMs)\cite{achiam2023gpt} are increasingly applied to robotic manipulation tasks, largely because of their extensive world knowledge gained from pretraining on vast datasets. However, due to VLM's pretraining being primarily based on text data and 2D image data from the Internet, and trained on user dialogue tasks, there is a capability between its reasoning abilities and the complex embodied tasks in the 3D real world. This gap makes it difficult for VLMs to fully meet the complex demands of real-world tasks, posing challenges for their application in real world robotic manipulation.

Recent research \cite{huang2024rekep,huang2024copa,pan2025omnimanip,tang2025geomanip} has introduced promising methods to bridge this gap. These approaches employ spatial representations (e.g., points or vectors) to represent scene entities and utilize VLMs reasoning to define constraints between these representations. Robotic tasks are executed through interactions between these constraints and dynamically tracked spatial representations. The main advantages of these methods include eliminating the need for expensive, high-quality robotic interaction data and providing strong generalization capabilities across different scenarios and robots.

However, existing approaches still face notable challenges. Their spatial representation extraction schemes are typically Fixed or Task-Independent, lacking the flexibility to dynamically adapt to the requirements of different tasks or different stages within a single task. Consequently, this can result in Insufficient representational ability or Prolonged extraction times. For instance, a simple task like "Grasp and Place Plush Cat" may only require extracting the object's centroid as a keypoint. In contrast, "Placing Plush Cat Upright and Facing the Camera" necessitates precisely extracting the object's 6D pose. Furthermore, if the task additionally demands orienting the cat’s tail toward a specific direction, an even finer-grained spatial representation, such as a vector indicating tail orientation, becomes necessary. These examples clearly show that the Complexity of a task determines the Types and Granularity of spatial representations required, and that Stronger representational ability typically entails Higher extraction costs. Therefore, methods relying on Fixed extraction schemes (e.g., DINO+SAM+clustering for keypoint extraction) struggle to accommodate varying task needs effectively, impacting task performance adversely. We define the problem as: How to provide Sufficient and Efficient spatial representation extraction support for VLM-guided robotic manipulation methods.

We propose T‑Rex, a Task‑Adaptive Spatial Representation Extraction approach, dynamically selects the optimal spatial representation extraction scheme for each entity in real-time based on the specific task and scene, effectively balancing representation capability and extraction efficiency. Additionally, to manage the complexity introduced by T-Rex, we introduce the Chain of Grounding (CoG) method, guiding the VLM through a robust, step-by-step grounding of instructions.

Specifically, we first construct an Extensible Spatial Representation Extraction Toolkit Framework integrating diverse Large Vision Models for spatial representation extraction. Through CoG, we explicitly define the VLM’s instruction grounding process. When the VLM determines that a task stage requires finer-grained spatial representations, it triggers local image extraction for the relevant region and subsequently extracts the Fine-Grained spatial representations. After obtaining the Heterogeneous and Multi-Granular spatial representations, the VLM will construct constraints based on the acquired Heterogeneous and Multi-Granular spatial representations. Then, Low Level Action Sequence Generator generate the robot action sequence based on the constraints and the tracked spatial representations.

Our method offers several key: (1) Comprehensive and Efficient spatial representation extraction: Due to the Task-Adaptive spatial representation extraction scheme, our approach simultaneously achieves powerful spatial understanding and high operational efficiency; (2) Extensibility: New spatial representation extraction tools can be easily integrated into the system; (3) Stability: The system can operate robustly and stably, including the adaptive extraction module described in (1); (4) Plug-and-play: It can be rapidly deployed without requiring Any training.

The primary contributions of our work include: (1) Proposing a Task-Adaptive spatial representation extraction method covering adaptive extraction of Heterogeneous and Multi-Granularity spatial representations; (2) Designing an extensible and open Spatial Representation Extraction Toolkit framework with practical examples provided; (3) Proposing CoG to explicitly guide VLM in step-by-step instruction grounding.

\section{Related Works}

\paragraph{Vision-Language Models for Robotics.}

Vision-Language Models (VLMs) are increasingly applied in robotic manipulation, attributable to their ability to acquire rich environmental understanding and high-level commonsense reasoning from large-scale pretraining data. Many existing studies focus on employing pretrained VLMs for task planning and high-level inference\cite{hu2023look,yang2024guiding,kumar2024open,gao2024physically,wang2024vlm}, while other approaches have begun to leverage inference capabilities of VLMs to guide more fine-grained robot actions beyond abstract planning\cite{qi2025sofar,zhao2024vlmpc,liu2025kuda,huang2023voxposer,liu2024moka}. Additionally, some works explore the use of VLMs for reward function design in reinforcement learning\cite{patel2025real}. Methods most relevant to our work typically first extract spatial representations of the scene (e.g., keypoints or vectors), then have the VLM generate constraints based on these representations, and finally use a solver to compute the robot action sequence\cite{huang2024rekep,pan2025omnimanip,tang2025geomanip}.

\paragraph{Grounding Language Instructions.}

Grounding Language Instructions has been extensively studied in robotic manipulation, aiming to map natural language instructions into forms that can relatively directly guide the robot in performing tasks. In earlier research, classical tools such as lexical analysis, formal logic, and graphical models were widely used to interpret and map language instructions\cite{thomason2015learning,kollar2014grounding,kollar2010toward,tellex2011understanding}. In recent years, end-to-end learning methods have gained significant application in robotic manipulation\cite{wen2024tinyvla,kim2024openvla,liu2024rdt,brohan2022rt,brohan2023rt,gu2023rt,aldaco2024aloha,fu2024mobile,li2024cogact,ze20243d,ze20243d,team2024octo,cheang2024gr,wang2025roboflamingo,li2023vision,liu2025hybridvla,zhao2025cot,li2025pointvla,qu2025spatialvla,shi2025hi,bjorck2025gr00t}. These methods implicitly ground language instructions by learning the mapping between language and actions from data with language annotations, although they face challenges such as lack of interpretability and high data requirements. In research on using Vision-Language Models (VLMs) for robotic manipulation, some methods attempt to leverage the powerful reasoning capabilities of VLMs to directly map natural language instructions to robot action sequences\cite{yang2025embodiedbench}. Other methods try to map language instructions to task planning representations such as text\cite{liu2023llm+} or PDDL\cite{song2023llm} . Additionally, some works explore mapping language instructions to executable code\cite{liang2023code,singh2023progprompt,vemprala2024chatgpt}. The line of work most similar to ours\cite{huang2024rekep,pan2025omnimanip,tang2025geomanip} maps language instructions to constraints, which then interact with the scene to guide robotic manipulation.

\paragraph{Vision Foundation Models and Spatial Representation for Robotics.}

Research on applying Vision Foundation Models(VFMs) to computer vision tasks is a extensive and active field. Robotic manipulation involves processing complex, real-world scene information; therefore, leveraging the visual comprehension capacity of VFMs for scene analysis presents a direct and effective approach. The application of VFMs to robotic manipulation has also become a thriving area of research. We refer readers to recent surveys for an overview of the latest advances in VFMs and their applications in robotics\cite{hu2023toward,awais2025foundation}. Existing work typically relies on VFMs to extract spatial representations from scenes—--using the visual comprehension capacity learned from large-scale image datasets to guide robotic manipulation—--because spatial representations offer a more immediate and effective approach of directing robotic actions compared to raw image inputs. Object detection \cite{wang2024yolov10,tian2025yolov12,wang2025yoloe,liu2024grounding}and segmentation models\cite{ravi2024sam,zhao2023fast,zou2023segment} can extract task-relevant bounding boxes or segmentation masks as spatial representations and have been widely adopted as 2D perception modules in robotic manipulation. Keypoints---valued for their flexibility, strong generalization, and ease of extraction—are among the most common spatial representations in robotic manipulation\cite{manuelli2019kpam,wen2023any,huang2024rekep,liu2024moka,nasiriany2024pivot}. Another common representation is the 6D pose: foundation models estimate an object’s full 6D pose via model-based\cite{wen2024foundationpose,lin2024sam,zhang2024omni6dpose} or model-free\cite{liu2022gen6d,sun2022onepose,he2022onepose++,cai2024gs} approaches, offering enhanced spatial representation capability to capture the object’s current pose. Several works have explored the utility of 6D poses as spatial representations for robotic manipulation tasks\cite{wen2022you}. Semantic vectors have been explored to encode task-relevant object orientations(e.g., camera-facing direction)\cite{qi2025sofar}. More recently, advances in 3D visual grounding have introduced 3D bounding boxes as a promising spatial representation\cite{li2024seeground,xu2024vlm,guo2025tsp3d,zhu2024scanreason}, enabling precise localization of objects’ positions and extents in three-dimensional space. 

In Appendix \ref{arw}, we discuss the Connections and Distinctions between our approach and existing work in these fields.

\section{Method}
\label{method}

\begin{figure}[tb]
  \centering
  \includegraphics[width=\linewidth]{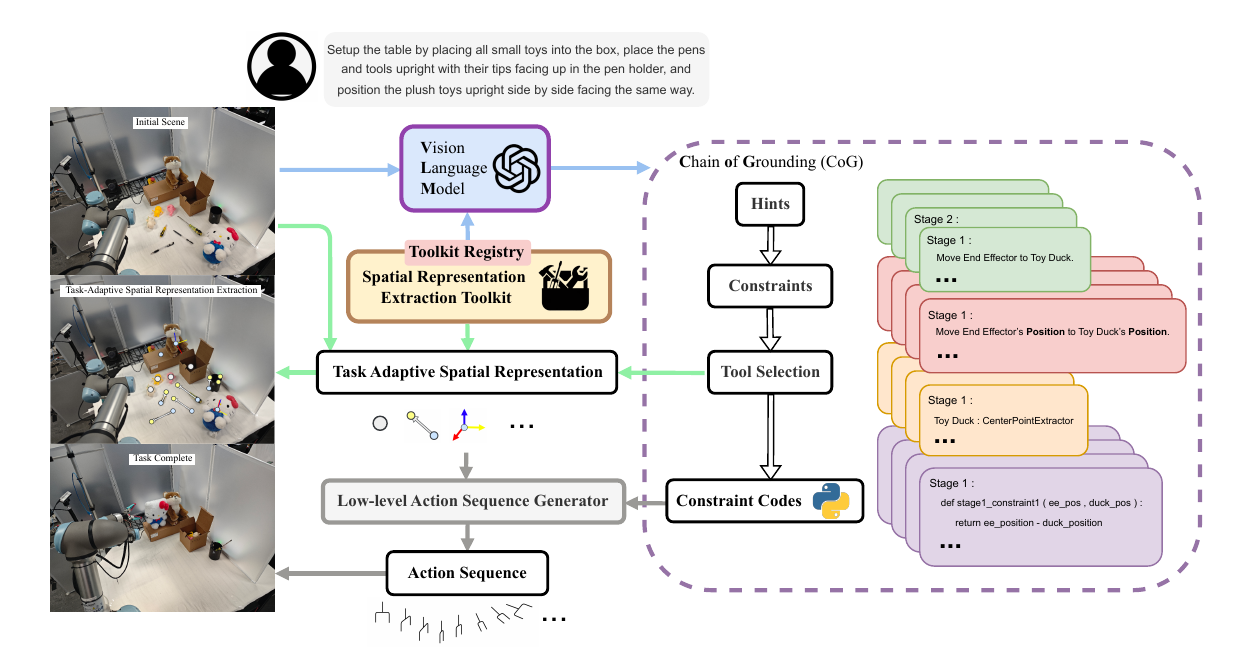}
  \caption{Overview framework. Given natural language instruction and scene observation, VLM guides system to adaptively invoke the optimal extractor from Spatial Representation Extraction Toolkit for each task-relevant object to derive spatial representations. Low-Level Action Sequence Generator then generates robot's action sequence based on these representations and VLM-inferred constraints to complete the task.}
  \label{fig1main}
\end{figure}

An overview of our approach is presented in Fig. \ref{fig1main}. We begin by providing a concise formalization of the robotic manipulation problem and our framework (Sec. \ref{3.1pf}), then give a detailed description of our Task-Adaptive, Heterogeneous Multi-Granularity representation extraction method (Sec. \ref{3.2ta}), and finally introduce the Chain of Grounding (CoG) module for guiding the VLM in robust reasoning over robotic manipulation tasks (Sec. \ref{3.3cog}).

\subsection{Problem Formulation}
\label{3.1pf}

Given a free-form natural-language instruction \(\ell\) (e.g., “pour tea into the cup”) and an initial scene observation \(X_{0}\) (e.g., an RGB-D image), our goal is to produce a sequence of end‑effector trajectories \(\{\tau_{i}\}_{i=1}^{N}\) corresponding to subtasks \(\{\ell_{i}\}_{i=1}^{N}\) into which \(\ell\) is decomposed. Each trajectory \(\tau_{i}\) is a sequence of waypoints
\[
\tau_{i}=\bigl\{(p_{i,t},\,g_{i,t})\bigr\}_{t=1}^{T_{i}},
\]
where \(p_{i,t}\in SE(3)\) is the desired 6-DoF end-effector pose at time \(t\), and \(g_{i,t}\in\{\text{open},\text{close},\text{hold}\}\) is the gripper command. Concatenating all \(\tau_{i}\) yields the full manipulation plan that transforms the initial scene \(X_{0}\) to satisfy the instruction \(\ell\).

\subsection{Task-Adaptive Spatial Representation Extraction}
\label{3.2ta}

Our Task-Adaptive Spatial Rpresentation Extraction Framework is composed of two core modules: the task-adaptive extraction of Heterogeneous spatial representations (Sec. \ref{321hete}) and the task-adaptive extraction of Multi-Granularity spatial representations (Sec. \ref{322mg}).

\subsubsection{Task-Adaptive Extraction of Heterogeneous Spatial Representations}
\label{321hete}

\begin{figure}[tb]
  \centering
  \includegraphics[width=\linewidth]{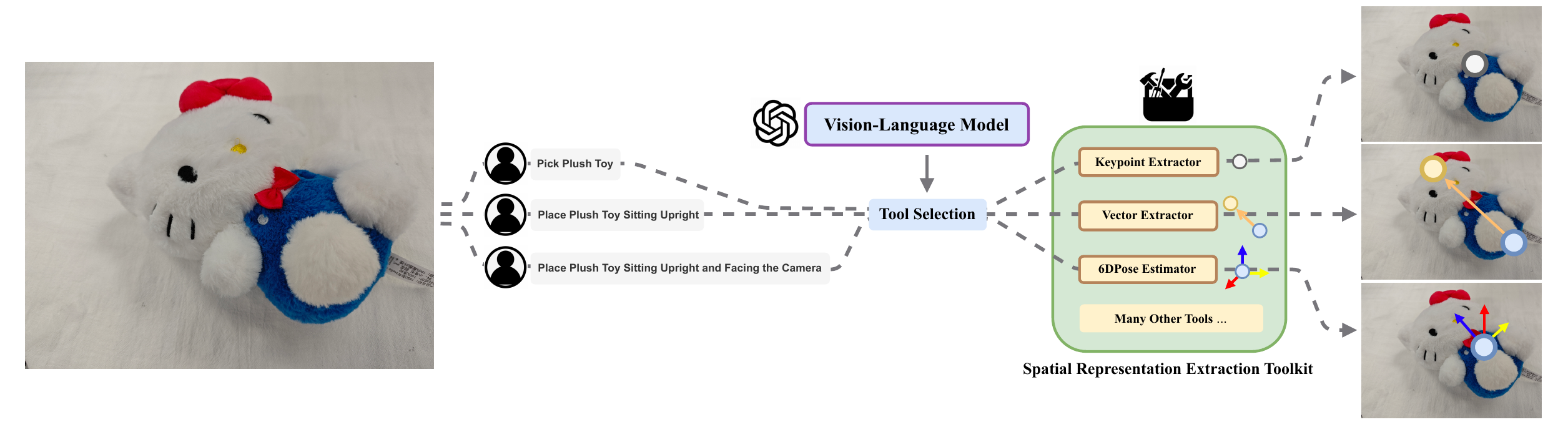}
  \caption{Spatial Representation Extraction Toolkit. System adaptively invokes the optimal extractor for each task to extract the required representation.}
  \label{fig:diagram}
\end{figure}

The Spatial Representation Extraction Toolkit includes a Registry containing information on each tool, maintained collaboratively by the system and users. The Registry is provided to the VLM at the first stage of the pipeline (see Sec. \ref{3.3cog} for details) as one of the inputs to CoG. We discuss the selection of tools in the Toolkit and show a sample registry in Appendix \ref{a61toolkitsamp}.

Formally, we represent the Spatial Representation Extraction Toolkit as a set of tools 
\(\mathcal{R}=\{T_i\}_{i=1}^N\), where each tool is defined in toolkit registry by
\[
T_i=(I_i,\;o_i,\;f_i,\;s_i,\;h_i),
\]
with \(I_i\) denoting the required inputs (e.g., RGB image, depth image, object list), \(o_i\) the spatial representation output (e.g., 6D pose, keypoints), \(f_i\) the output format (e.g., a \(4\times4\) SE(3) matrix), \(s_i\) a brief implementation summary, and \(h_i\) the historical average execution time maintained and updated after each invocation.  Given a natural-language instruction \(I\) and a scene observation \(X\), tasks are decomposed into a set of stages \(S=\{1,\dots,S\}\), each stage \(s\) having a set of task-relevant objects \(O_s\).  For each \(s\in S\) and \(o\in O_s\), the VLM selects
\[
t^*_{s,o}
=\arg\max_{t\in\mathcal{R}}\Bigl[P_{\mathrm{succ}}(t\mid I,X,s,o)\;-\;\lambda\,h_{t}\Bigr],
\]
where \(P_{\mathrm{succ}}(t\mid I,X,s,o)\) estimates the success probability of extracting the required spatial representation, \(\lambda\) trades off success likelihood against extraction cost \(h_t\), and \(t^*_{s,o}\) denotes the optimal tool selected for stage \(s\) and task-relevant object \(o\) under this criterion. To guide the VLM in effectively implementing this tool selection process, we detail the Chain of Grounding (CoG) in Sec. \ref{3.3cog}.

After selecting the most appropriate spatial representation extraction tool for each task-relevant object at each stage, we use the chosen tool to extract the spatial representation in that stage: 
\[
r_{s,o} = t^*_{s,o}(I, X),
\]
where \(r_{s,o}\) is the extracted spatial representation (e.g., keypoints or 6D pose), and the task scene observation \(X=\{c_{\mathrm{rgb}}, c_{\mathrm{depth}}\}\), with \(c_{\mathrm{rgb}}\) the RGB image from the camera and \(c_{\mathrm{depth}}\) the corresponding depth map.

In the first stage of the framework, the VLM uses CoG to generate a series of Spatial Constraints represented as Python functions; each constraint function takes the Spatial Representations extracted by the tools selected from the toolkit as input and returns a Scalar Cost that measures how well the constraint is satisfied---when a constraint is met, its cost approaches or equals zero. The details of Constraints based on Heterogeneous Representations are presented in Appendix \ref{a3con}.

In the third stage of the framework, we enter Low-Level Robot Action-Sequence Generation. For common constraints built on typical spatial representations (e.g., points, vectors, or 6D poses)---whose cost outputs are usually continuous---we use a solver to directly compute the robot's action sequence. The solver takes as input all of this stage's constraint functions, the tracked spatial representations, and other fixed constraints (such as collision avoidance and IK based reachability constraints). This component follows many of the practices introduced in \cite{huang2024rekep}, we refer readers to \cite{huang2024rekep} for further details on constraint generation and solve.

In task stages where constraints rely on unconventional representations, we employ an alternative approach to generate the robot's action sequence. Following practices from\cite{liang2023code}, we leverage the VLM to assemble a complete Python policy script in which these constraint functions participate(e.g., within control flow logic) to shape the robot's behavior and ensure correct execution. Further details on representations (such as state machines or topologies) are provided in Appendix \ref{a61toolkitsamp}.

The toolkit's key function lies in enabling Task-Adaptive Efficiency-Accuracy Trade-Offs; at the same time, its open Registry supports user customization-simply by specifying a few parameters in a configuration file, new representation-extraction tools can be seamlessly integrated.

\subsubsection{Task-Adaptive Extraction of Multi-Granularity Spatial Representations}
\label{322mg}

\begin{figure}[tb]
  \centering
  \includegraphics[width=\linewidth]{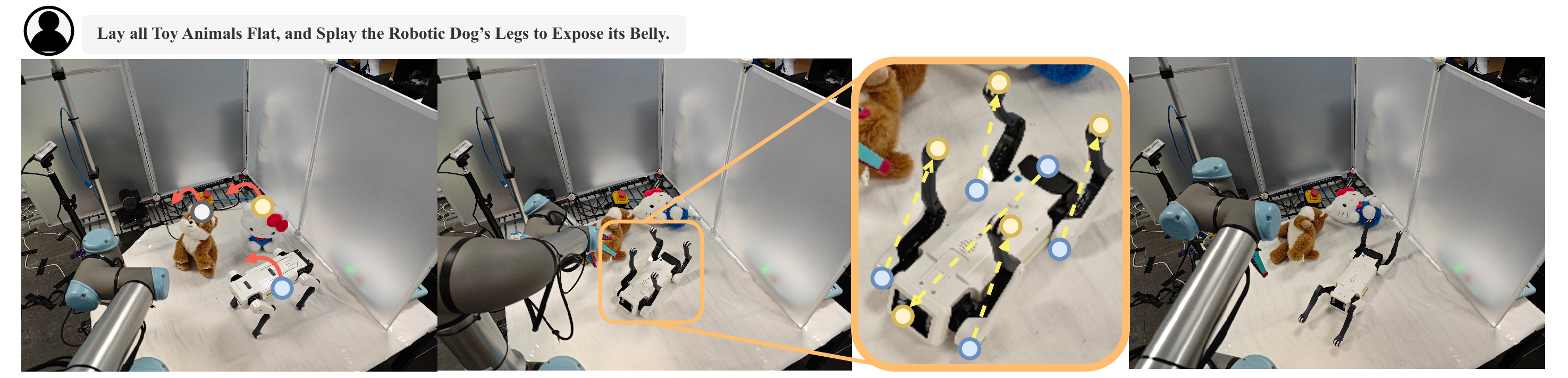}
  \caption{Multi-Granularity spatial representation extraction. First crop a local subgraph, then extract representations from this subgraph.}
  \label{figmg}
\end{figure}

In certain manipulation tasks (e.g., requiring all toy animals on the table to be laid flat, and in the case of a robotic dog, splay its legs to expose its belly) Fine-Grained spatial representations extracted from the raw camera view alone often proved insufficient. When a task demands precise manipulation of a local region of an object (e.g., the legs of the robotic dog), existing extractors we have tried frequently fail to reliably and accurately capture the necessary Fine-Grained spatial representations(e.g., the orientation of the robotic dog's legs), as evidenced by our ablation studies in Sec. \ref{43abl}.

We propose a Multi‑Granularity Spatial Representation Extraction strategy, a lightweight, attention‑inspired local zoom strategy, as illustrated in Fig. \ref{figmg}. By default, we extract coarse-grained spatial representations over the complete RGB frame without any additional preprocessing (see Sec. \ref{321hete} for details). If, during the first‑stage VLM invocation in the pipeline, the VLM judges that a given spatial representation at stage s requires finer detail (e.g., the leg of the robotic dog), a preprocessing step is invoked prior to spatial representations extraction. First, a target segmentation module (e.g., SAM) isolates the object to localize the relevant object region; this mask is then expanded by a fixed or adaptive padding and used to crop a local sub‑image. Next, the adaptive extraction method described in Sec. \ref{321hete} is applied to this sub‑image. Finally, the resulting Fine‑Grained representations are Merged Back into the main pipeline to guide subsequent robot action sequencing.

By employing this strategy, we explicitly restrict downstream extraction modules to the correct target regions, thereby yielding higher‑quality fine‑grained representations. Moreover, this approach is Task‑Adaptive, it activates only when necessary.

\subsection{Chain of Grounding (CoG)}
\label{3.3cog}


The Task-Adaptive Heterogeneous and Multi-Granularity spatial representation extraction methods discussed in Sec. \ref{3.2ta}, while enhancing the system's spatial understanding and preserving operational efficiency, impose an additional reasoning burden on the VLM, forcing it to perform task decomposition, constraint code generation, and representation extraction tool selection all within a single call. As shown by our experiments (see Sec. \ref{43abl}), without any explicit guidance the VLM frequently produces incomplete or incorrect constraints or chooses clearly unsuitable extraction tools, leading to task failures or unacceptably long processing times. To address this, we introduce and apply the Chain of Grounding (CoG) method to the VLM at the front of the framework's execution pipeline, explicitly guiding the VLM's reasoning to improve stability.

Our key insight is that grounding a natural-language instruction and scene observation in our framework naturally decomposes into four sequential, dependent phases: operational hint inference, constraint inference, tool selection, and constraint code generation. Accordingly, we designed the Chain of Grounding (CoG) to explicitly navigate the VLM through these stages. First, CoG decomposes the task into multiple stages and generates a diverse set of concise, representation-agnostic operational hints. These hints are brief reminders of critical considerations during task execution, aimed at ensuring the successful completion of the task. Next, for each tip, CoG infers the Spatial Constraints required at that stage in natural-language form, grounded in the scene's objects and their spatial representations. It then queries the Toolkit Registry to select the optimal extraction tool for every Task-Relevant object in each stage, balancing representational capacity against extraction efficiency. Finally, CoG converts the natural-language constraints into executable Python functions, each consuming the corresponding spatial representations and returning a Scalar Cost that quantifies constraint satisfaction. Collectively, CoG outputs all per-stage Python constraint functions and the chosen spatial representation extraction tools for every task-relevant object.

Briefly, CoG is defined by the mapping
\[
\{(\{T_{s,o}\}_{o\in O_s},\,F_s)\}_{s=1}^S = \mathscr{G}(I, X),
\]
which grounds a natural‐language instruction \(I\) and scene observation \(X\) into per‐stage tool selections \(T_{s,o}\) for each task‐relevant object and a series of executable constraint functions \(F_s\) in every stage. We only provide a high‐level formalization of the Chain of Grounding (CoG) in main text, detailed formalization of the decomposition, constraint generation, tool selection, and code generation are provided in Appendix \ref{a52cog}.

\section{Experiments}
\label{headings}

In this section, we seek to answer the following research questions: (1) How effective and efficient is our system on open-vocabulary manipulation tasks across diverse real-world scenarios? (Sec. \ref{4.1db}); (2) How does each components contribute to the overall system performance?  (Sec. \ref{43abl}); (3) To what extent do individual system components contribute to failures, and what are the primary sources and failure modes? (Sec. \ref{44error}).

To validate T-Rex, we constructed a real-world desktop platform: an Intel RealSense D435i RGB-D camera was calibrated and mounted at an optimal overhead position for scene perception, and a UR5e 6-DoF arm with a 1-DoF gripper executed the manipulation tasks. Further hardware and experimental setup details are provided in Appendix \ref{a1real}. For the Vision-Language Model, we employ GPT-4.1 for inference. In Appendix \ref{apvlm}, we analyze the impact of different VLM choices on the system's overall performance.\warn{setup 1 duan}

\subsection{Real-World and Open-Vocabulary Manipulation}
\label{4.1db}

We designed and selected 15 real-world, open-vocabulary manipulation tasks ranging from simple placement operations to challenging scenarios involving complex spatial constraints. Appendix\ref{a1real} contains extended details on task definitions, baselines, etc. For each task, we conducted 10 independent trials with the poses of task-relevant objects randomized in every trial, and randomized both the location and orientation for every object relevant to the task. The evaluation metrics include task execution success rate and task execution time. We compared our method against VoxPoser and ReKep as baselines. 

\begin{figure}[tb]
  \centering
  \includegraphics[width=\linewidth]{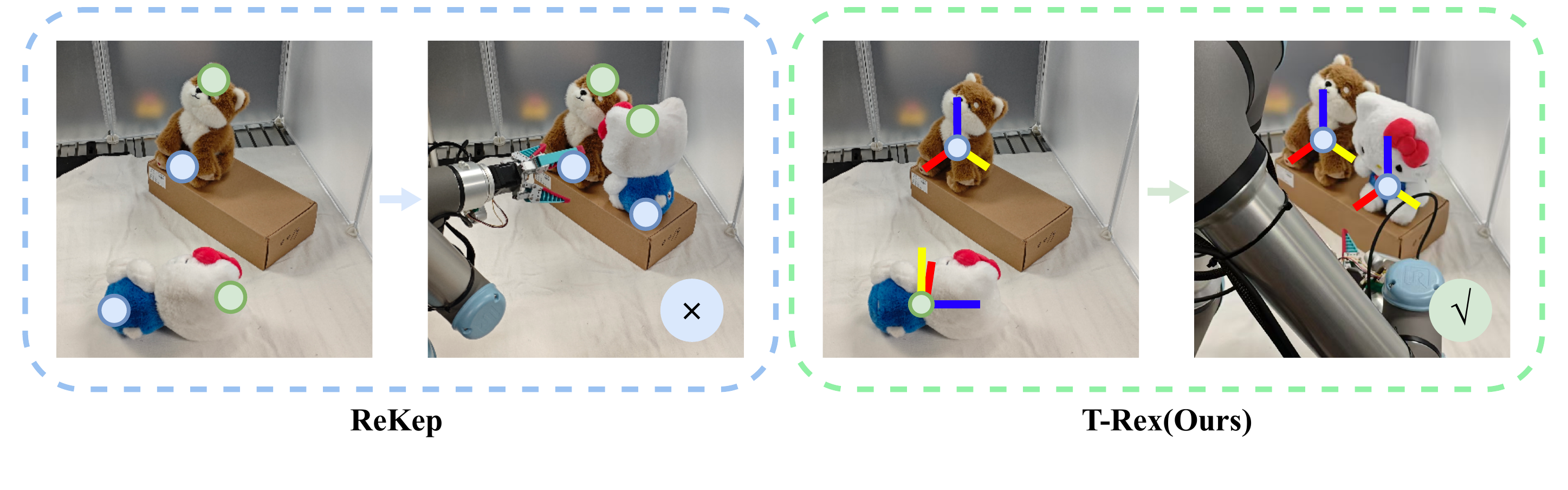}
  \caption{Because T-Rex Toolkit includes 6D pose estimation module, it outperforms the keypoint-only baseline ReKep on tasks that demand object pose estimation.}
  \label{figcomp}
\end{figure}

Tab. \ref{t1db} presents the detailed quantitative results. Our method demonstrates robust zero-shot generalization and comprehensive spatial understanding, achieving significant performance gains over the baseline approaches. We observe that, compared to methods employing a fixed pipeline to extract a single spatial representation, our Task-Adaptive paradigm---which selects the most appropriate representation extraction scheme for each object at each stage---is far more flexible. This design endows our system with both excellent spatial understanding (e.g., correctly orienting plush toys upright in the same direction as shown in Fig. \ref{figcomp}) and comparatively favorable runtime efficiency, consistent with the analysis in Sec. \ref{3.2ta}. Furthermore, our approach yields a higher average success rate on multi-stage tasks. This improvement can be attributed to the explicit guidance provided by CoG, which effectively mitigates VLM hallucinations, fully leverages the VLM’s reasoning capabilities, and robustly grounds open-vocabulary instructions into comprehensive and correct constraints as well as spatial representations for every object at each stage. The critical importance of CoG within our framework will be examined in detail in the ablation study of Sec. \ref{43abl}.

Although our experiments clearly demonstrate the advantages of our approach, we must acknowledge several limitations. In Sec. \ref{5conclu}, we discuss the key limitations of our method.

\begin{table}[tb]
  \centering
  \caption{Quantitative results in real-world experiments. T-Rex achieves high success rates on everyday manipulation tasks, demonstrating superior spatial understanding and overall execution efficiency, and significantly surpassing the baseline VoxPoser and ReKep.}
  \label{t1db}
  \begin{tabular}{l cc cc cc}
    \toprule
    \multirow{2}{*}{\textbf{Task}} 
      & \multicolumn{2}{c}{\textbf{VoxPoser}} 
      & \multicolumn{2}{c}{\textbf{ReKep}(Auto)} 
      & \multicolumn{2}{c}{\textbf{T-Rex (Ours)}} \\
    \cmidrule(lr){2-3} \cmidrule(lr){4-5} \cmidrule(lr){6-7}
    & Success & Time(s) & Success & Time(s) & Success & Time(s) \\
    \midrule
    Open Drawer
      & 4/10  & 17.5 
      & 2/10  & 18.1 
      & 6/10  & 14.3 \\
    Open Bottle
      & 7/10  & 13.3 
      & 7/10  & 19.8 
      & 8/10  & 13.5 \\
    Sweep Trash
      & 9/10  & 11.4 
      & 8/10  & 11.8 
      & 9/10  & 12.9 \\
    Pack Shoes
      & 0/10  & 37.4 
      & 2/10  & 44.6 
      & 5/10  & 38.8 \\
    Pour Water
      & 0/10  & 25.7 
      & 3/10  & 27.5
      & 7/10  & 24.1 \\
    Recycle Can
      & 6/10  & 19.3 
      & 7/10  & 21.9 
      & 9/10  & 14.2 \\
    Fold Coat
      & 0/10  & 40.9 
      & 4/10  & 47.7 
      & 4/10  & 38.8 \\
    Stack Block
      & 6/10  & 90.7
      & 6/10  & 101.8 
      & 6/10  & 74.5 \\
    Hammer the Button
      & 0/10  & 27.1 
      & 1/10  & 27.4 
      & 4/10  & 29.4 \\
    Close Lid of Laptop
      & 4/10  & 19.8 
      & 2/10  & 23.1 
      & 7/10  & 21.6 \\
    Setup table: Toys Place  
      & 5/10  & 68.4 
      & 5/10  & 70.4 
      & 7/10  & 54.6 \\
    Setup table: Tools Insert
      & 0/10  & 52.9 
      & 3/10  & 57.6 
      & 7/10  & 56.3 \\
    Setup table: Plush Toys Place
      & 1/10  & 31.5 
      & 1/10  & 33.3 
      & 4/10  & 26.9 \\
    Setup table: Mixed 
      & 0/10  & 254.9 
      & 0/10  & 240.9 
      & 2/10  & 217.5 \\
    \midrule
    \textbf{Total}
      & 30\% & 50.8
      & 36.4\% & 53.3 
      & \textbf{60.7\%} & \textbf{45.5} \\
    \bottomrule
  \end{tabular}
\end{table}

\subsection{Ablation Study}
\label{43abl}

To quantitatively evaluate how our two core modules, Chain of Grounding (CoG) and the Spatial Representation Extraction Toolkit, each contribute to overall system performance, we conducted a set of ablation studies. These ablations reuse the tasks from Sec. \ref{4.1db} and evaluate success rate and average completion time as metrics. When the Toolkit is removed, we employ a single fixed extractor; when CoG is removed, the VLM is invoked with only a simple prompt to produce the required outputs.

\begin{figure}[tb]
  \centering
  \begin{minipage}[t]{0.5\textwidth}
    \vspace{0pt}
    \captionof{table}{Ablation study results. SP denotes the simplest point extractor (Grounding DINO + center point); VPV denotes the VLM-based point and vector extractor. Results are reported over 3 runs different seeds.}
    \label{tab:abla}
    \begin{tabular}{@{} l @{\hskip 4pt} c @{\hskip 4pt} c @{}}
      \toprule
      \textbf{Configuration} & \textbf{Success(\%)} & \textbf{Time (s)} \\
      \midrule
      Ours            & $60.7\%_{\scriptstyle\pm2.1\%}$                        & $45.5_{\scriptstyle\pm1.3}$                     \\
      w/o CoG           & $52.1\%_{\scriptstyle\pm2.4\%}$                        & $41.4_{\scriptstyle\pm2.1}$                     \\
      w/o Toolkit (SP)  & $30.7\%_{\scriptstyle\pm3.7\%}$                        & $33.6_{\scriptstyle\pm3.6}$                     \\
      w/o Toolkit (VPV) & $55.0\%_{\scriptstyle\pm2.9\%}$                        & $47.9_{\scriptstyle\pm3.2}$                     \\
      w/o CoG, w/o Toolkit (SP) & $27.9\%_{\scriptstyle\pm1.5\%}$               & $30.0_{\scriptstyle\pm0.9}$                     \\
      w/o CoG, w/o Toolkit (VPV)& $51.4\%_{\scriptstyle\pm1.4\%}$              & $44.3_{\scriptstyle\pm4.4}$                     \\
      \bottomrule
    \end{tabular}
  \end{minipage}
  \hfill
  \begin{minipage}[t]{0.4\textwidth}
    \vspace{20pt}
    \includegraphics[width=\linewidth]{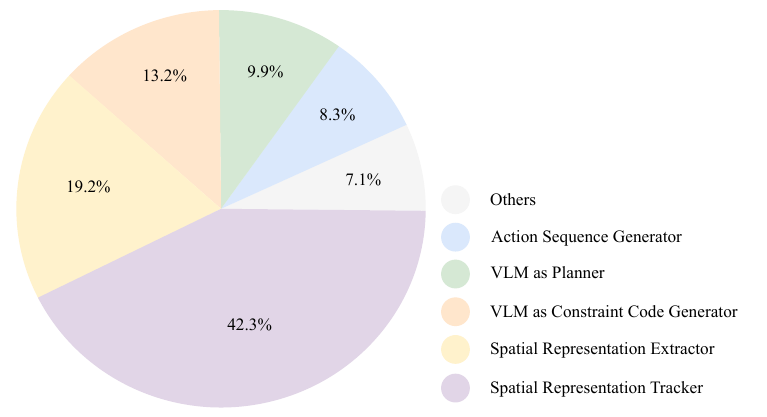}
    \caption{System error breakdown.\warn{zi too small}}
    \label{fig:error}
  \end{minipage}
\end{figure}

The ablation results are summarized in Tab. \ref{tab:abla}. We first observe that integrating CoG yields a clear improvement in success rate with insignificant additional latency. Although the improvement is modest, CoG consistently delivers a stable uplift, as a simple yet effective prompt-engineering mechanism, validates our design and understanding of the instruction grounding process (see Sec. \ref{3.3cog}). Conversely, when the Toolkit is removed and a fixed spatial representation extractor is used, the system must trade off success rate against overall execution efficiency, which is consistent with our analysis in Sec. \ref{3.2ta}.

\subsection{System Error Breakdown}
\label{44error}

Due to the framework’s interpretability and modular design, we can precisely identify and analyze the sources of system failures, as shown in Fig. \ref{fig:error}. With the inclusion of CoG, errors from the planning module (VLM) are effectively negligible, consistent with our experimental evidence and validating the robustness of the instruction-grounding process. While the Spatial Representation Extraction Toolkit is a substantial component, yet it maintains a low error rate relative to its complexity. We must acknowledge that this is due to the stability of the sample tools we selected.

Across all modules, Spatial Representation Tracking introduces the most significant errors. Most existing extraction tools lack continuous tracking capabilities and are not optimized for rapid motion or intermittent occlusions, making consistent representation challenging under these conditions. This gap indicates that robust spatial tracking slightly exceeds the capabilities of current tools. We discuss this in Appendix \ref{a64track}

Furthermore, we observe that employing a more powerful VLM markedly improves performance in the constraint generation stage—this finding aligns with our experimental results and highlights the high inference demands of this stage (see Appendix \ref{apvlm}). Finally, while the low level action sequence generator and other modules can also cause occasional failures, their impact on overall system stability is relatively minor compared to the modules discussed above.

\section{Conclusion}
\label{5conclu}

In this work, we present T-Rex, a Task-Adaptive Spatial Representation Extraction Framework that dynamically selects the optimal extraction scheme for each entity, thereby supporting the subsequent generation of spatial constraints and the solving of robotic action sequences. T-Rex achieves both powerful spatial representation capability and efficient extraction, which in turn endows the overall system with comprehensive spatial understanding and high execution efficiency. The Chain of Grounding (CoG) method is applied to guide the VLM through progressive instruction grounding, ensuring stability throughout the reasoning process. Based on an extensible Toolkit Framework, T-Rex flexibly integrates various Large Vision Models and exhibits excellent system scalability. Notably, all without any additional training. Extensive experiments in both simulated and real-world robotic environments demonstrate that our approach offers significant advantages in spatial understanding, efficiency, and system stability. While advantageous, T-Rex also has limitations. In Appendix \ref{limi}, we discuss the limitations of T-Rex and outline several directions for future work.

\bibliographystyle{unsrt}
\bibliography{reference}

\newpage

\appendix

\section{Technical Appendices and Supplementary Material}

\subsection{Real-World Environment Setup}
\label{a1real}

\subsubsection{Hardware Setup}
\label{a11hard}

\begin{figure}[ht]
  \centering
  \includegraphics[width=\linewidth]{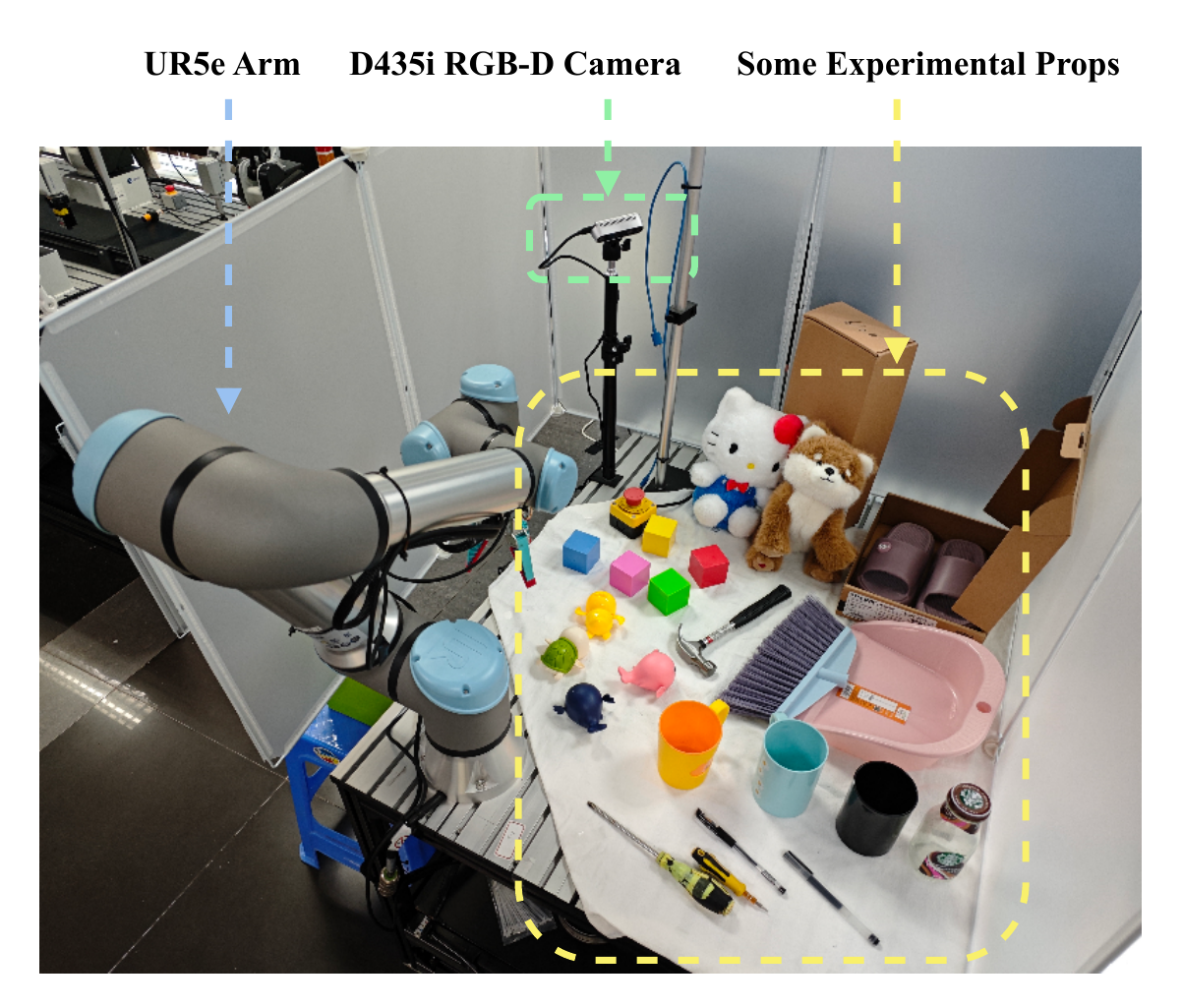}
  \caption{Single-Arm Platform.}
  \label{figUR}
\end{figure}

As shown in Fig. \ref{figUR}, we deploy a UR5e 6-DoF arm on our desktop platform, outfitted with a stepper-motor-driven 1-DoF gripper for object grasping. Robot control from the host PC is implemented via URScript in conjunction with the RTDE interface, supporting a theoretical maximum communication rate of 500 Hz. Because our method requires continuous and smooth end-effector rotation, we 
use a quaternion (\(\mathrm{quat}\)) representation rather than traditional 
Euler angles (\(\mathrm{rx}\!-\!\mathrm{ry}\!-\!\mathrm{rz}\)) to avoid the gimbal-lock singularities and abrupt discontinuities inherent to Euler formulations. Scene perception relies solely on the depth and color streams of a single Intel RealSense D435i RGB-D camera, demonstrating the low sensor-hardware requirements of our approach in practical deployment.
Our experiments are conducted on an Intel(R) Core(TM) i7-14700KF CPU and an NVIDIA RTX A6000 GPU.

\subsubsection{Tasks}
\label{a12task}

\begin{figure}[ht]
  \centering
  \includegraphics[width=\linewidth]{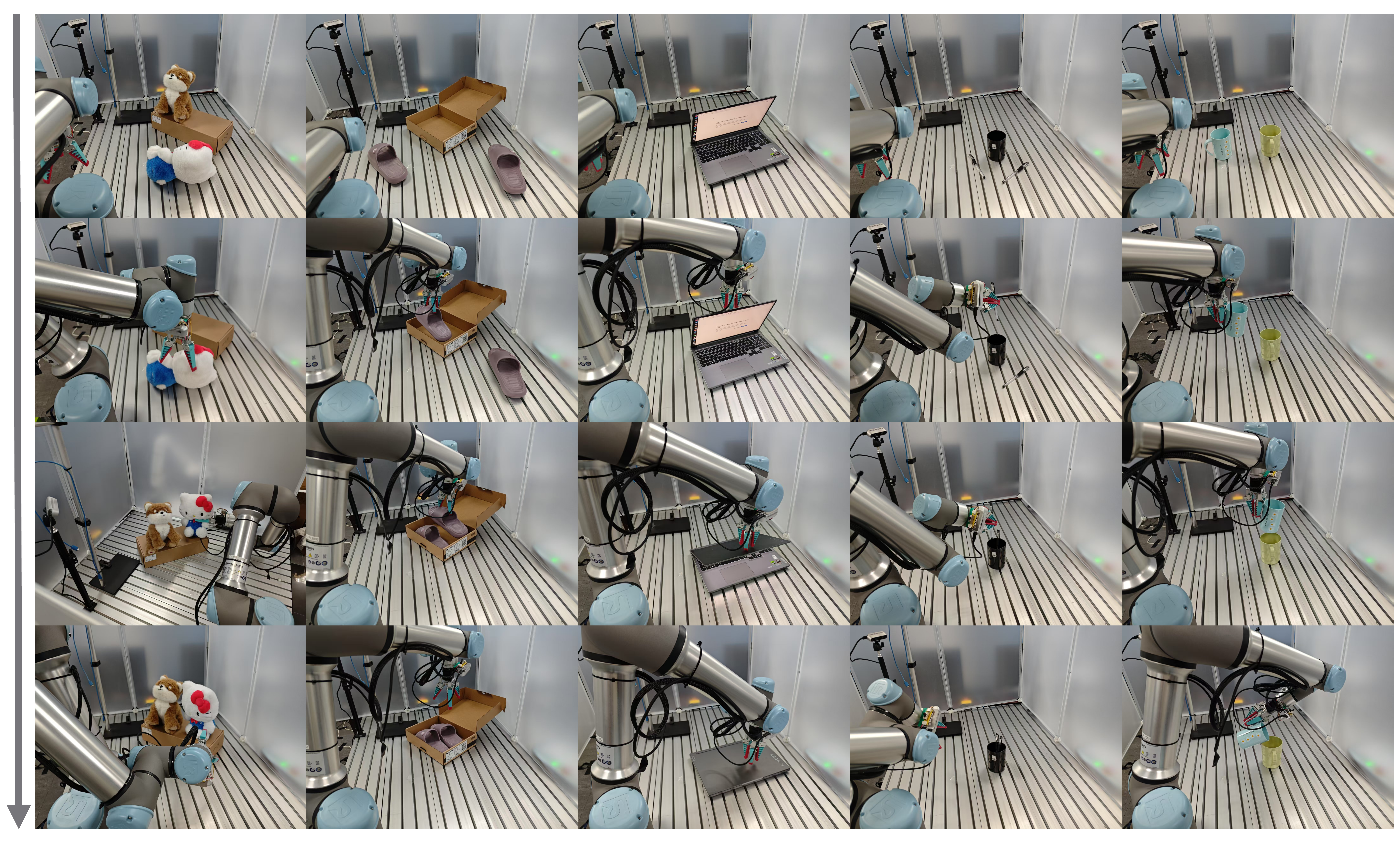}
  \caption{Execution Snapshots for Sample Tasks}
  \label{figExpSamp}
\end{figure}

In our task design, some tasks are adopted from related works\cite{huang2024rekep}\cite{huang2024copa}\cite{huang2023voxposer} while others are newly created to showcase the strengths of our method. Overall, the tasks are drawn from everyday life and are organized with a clear progression from simple to challenging. Fig. \ref{figExpSamp} provides snapshots of the execution process for some tasks.

\textbf{Open drawer:} The experimental setup consists of a small drawer placed on a table. The drawer's position and orientation are randomized to some extent but kept within the robotic arm's workspace. The task is divided into two stages: grasp the drawer handle and pull it outward. This requires correctly identifying and gripping the drawer handle, then pulling it in the proper direction. The success criterion is that the drawer is pulled out by at least one-third of its depth.

\textbf{Open Bottle: }The experimental setup includes a glass bottle with a cap standing upright on the table; its position are randomly varied within the robotic arm’s workspace. The task is divided into two stages: grasp the cap and lift it straight up to remove it. This requires correctly identifying the cap's three-dimensional position and applying a vertical pulling motion. The success criterion is that the cap is fully removed without any liquid spilling from the bottle.

\textbf{Sweep Trash: }The experimental setup includes crumpled paper randomly scattered within the robot's workspace, plus a broom and a dustpan. The task is divided into two stages: position the broom along the line between the trash and the dustpan, then sweep the paper into the dustpan. This requires accurately locating both the paper and the dustpan, and using the broom to push the paper in the correct direction until it enters the dustpan. The success criterion is that all paper pieces end up inside the dustpan.

\textbf{Pack Shoes: }The experimental setup includes one pair of slippers and an open cardboard box randomly placed within the robot's workspace. The task consists of four stages: grasp each slipper in turn and place it into the box. This requires detecting each slipper's orientation and placing it into the corresponding spot in the box, which is just large enough to hold both slippers. The success criterion is that both slippers are fully inside the box, do not overlap, and have their uppers facing upward.

\textbf{Pour Water: }The experimental setup includes two identical-model plastic cups of different colors randomly placed within the robot's workspace (Cup A contains water; Cup B is empty). The task requires correctly identifying and aligning Cup A's opening toward Cup B and controlling the tilt to pour. The success criterion is that water transfers into Cup B without any residual drops on the table or on Cup A.

\textbf{Recycle Can: }The experimental setup includes an empty beverage can and a two-compartment bin labeled "Recyclable" and "Non-Recyclable", both placed at random within the robot's workspace. The task is divided into two stages: grasp the can and place it into the "Recyclable" compartment. This requires correctly classifying the can as recyclable and locating the bin's compartments. The success criterion is that the can rests inside the "Recyclable" slot without tipping over the bin.

\textbf{Stack Block: }The experimental setup includes 3–6 identical cubes (edge length 5 cm) randomly scattered within the robot's workspace. The task is to pick up each cube from the table in sequence and stack them vertically at a predefined placement point (e.g., on a red reference block). This requires accurately detecting the positions of both the cubes and the stacking location, and planning precise grasp-and-place trajectories to form a stable column.

\textbf{Fold Coat: }The experimental setup includes a coat laid flat on the table. The task requires the robot to repeatedly grasp one end of the coat, pull it toward the opposite end, and lay it down to incrementally fold the coat. The success criterion is that the folded coat occupies less than half of its original tabletop area.

\textbf{Hammer the Button: }The experimental setup includes a hammer and a red push-button randomly placed within the robot's workspace. The task is divided into three stages: grasp the hammer's handle, orient the hammer head toward the button, and strike the button with the hammer head. This requires precise recognition and control of the hammer head's orientation and its relative position to the button. The success criterion is that the hammer head contacts the button in a downward orientation.

\textbf{Close the Lid of the Laptop: }The experimental setup includes a laptop with its screen fully opened placed on the table; its position and orientation are randomized within the robot's workspace. The task is divided into two stages: move the end effector to the upper edge of the screen's back and then smoothly press down along the hinge until the lid is fully closed. This requires correctly identifying the screen's edge and the hinge direction. The success criterion is that the lid sits flush against the keyboard with no gaps.

\textbf{Setup table - Toys Place: }Quickly Place All Toys in Box. The experimental setup includes various toys randomly scattered on the table and an open-top box, all within the robot's workspace. The success criterion is that all toys are transferred into the box within 60 seconds.

\textbf{Setup Table - Tools Insert: }Put All the Tools in the Pen Container with Handles Facing Up. The experimental setup includes multiple hand tools randomly scattered on the table and a pen holder beside them, all within the robot's workspace. The task requires recognizing each tool's handle end and placing the tool into the holder so that every handle faces upward. The success criterion is that all tools rest in the holder, handles oriented up, without tipping it over.

\textbf{Setup Table - Plush Toys Place:}Place All Plush Toys into the Box Upright and Facing the Same Direction. The experimental setup includes plush toys randomly scattered on the table and an open-top storage box, all within the robot's workspace. The task requires identifying each toy's front face and bottom, then placing them upright in the box so they all face the same way. The success criterion is that all plush toys stand securely in the box with their fronts aligned.

\textbf{Setup Table - Mixed:}  The experimental setup includes a mix of plastic toys, tools, and plush toys randomly scattered on the table, plus two open-top boxes and a pen holder, all within the robot's workspace. The robot must complete the three "Setup table" subtasks above in sequence.

\subsubsection{Baseline Methods}

\paragraph{VoxPoser}
VoxPoser\cite{huang2023voxposer} uses Vision–Language Models (VLMs) and Large Language models (LLMs) for reasoning. It infers affordances from natural-language instructions and then maps them into composable 3D value maps, which serve as objective functions for the planner to enable zero-shot trajectory synthesis.

\paragraph{ReKep}
ReKep\cite{huang2024rekep} (\emph{Relational Keypoint Constraints}) formulates a manipulation task as a set of Python functions that assign numerical costs to predefined 3D semantic keypoints in the scene, and then uses hierarchical optimization to solve for continuous end-effector actions in SE(3) in real time. "Auto" in Tab.\ref{t1db} denotes automatic generation of the Python functions by the VLM, as opposed to human annotation.

\paragraph{Experimental Configuration Consistency}
In our experiments, all methods invoke the same Vision–Language Model versions and, likewise, we keep other configurations such as the low-level control interfaces described in \ref{a11hard} as consistent as possible.

\subsubsection{Visuals Illustrations}

\subsection{Details of Constraints}
\label{a3con}



\begin{lstlisting}[
  float,                   
  floatplacement=!htbp,    
  caption={Example of point-based constraints.},
  label={lst:mdinput},
  language=Python
]
def stage1_constraint1(end_effector, points):
    """Align the end-effector with the red block center (points[0])."""
    red = np.array([points[0]["x"], points[0]["y"], points[0]["z"]])
    cost = np.linalg.norm(end_effector - red)
    return cost

def stage2_constraint1(end_effector, points):
    """Position the red block above the green block (points[1]), with a fixed vertical offset (e.g., block height)."""
    green = np.array([points[1]["x"], points[1]["y"], points[1]["z"]])
    height_offset = np.array([0.0, 0.0, 0.05]) 
    target_pos = green + height_offset
    horizontal_cost = np.linalg.norm(end_effector[:2] - target_pos[:2])
    vertical_cost   = abs(end_effector[2] - target_pos[2])
    return horizontal_cost + vertical_cost

\end{lstlisting}

\begin{lstlisting}[
  float,                   
  floatplacement=!htbp,    
  caption={Example of pose-based constraints.},
  label={lst:mdinput},
  language=Python
]
def stage1_constraint1(end_effector, poses):
    """Align the end-effector with the teddy cat (pose index 0)."""
    cost = np.linalg.norm(end_effector - poses[0][:3, 3])
    return cost

def stage2_constraint1(end_effector, poses):
    """Position the teddy cat next to the teddy bear (pose index 1) and facing the same direction."""
    offset = np.array([0, -0.2, 0])  
    offsetted_position = poses[1][:3, 3] + offset
    position_cost = np.linalg.norm(poses[0][:3, 3] - offsetted_position)
    
    bear_orientation = poses[1][:3, :3]
    cat_orientation = poses[0][:3, :3]
    orientation_cost = np.linalg.norm(bear_orientation - cat_orientation)
    
    return position_cost + orientation_cost

def stage2_constraint2(end_effector, poses):
    """The robot must still be grasping the teddy cat."""
    cost = np.linalg.norm(end_effector - poses[0][:3, 3])
    return cost

\end{lstlisting}

We present two two constraint examples for basic tasks to help readers understand the constraint component of our method. Following best practices from related work\cite{huang2024rekep}, each constraint is implemented as a Python function that takes as input the spatial representations extracted by the selected tools (e.g., points or 6D poses) and outputs a scalar cost to guide the low-level action sequence generator. The first task---"grasp the red block and place it on the green block"---can be completed using only point-based constraints. The second task-"place the plush cat next to the plush bear while maintaining the same orientation"---requires 6Dpose-based constraints. By analogy, a more complex task may require multiple types of constraints to work together at the same time.

\subsection{Comparison of VLMs}
\label{apvlm}

\begin{table}[ht]
  \centering
  \caption{Exploratory study of VLM choice effects on system performance}
  \label{tabVLM}
  \begin{tabular}{lcc}
    \toprule
    \textbf{Model}               & \textbf{Inference Validity Rate (\%)} & \textbf{Total Success Rate (\%)} \\
    \midrule
    o3                           & 91.4                                & 61.4                              \\
    GPT-4.1                      & 90.0                                & 60.7                              \\
    GPT-4o mini                  & 85.7                                & 57.1                              \\
    gpt-4-vision-preview         & 83.8                                & 55.0                              \\
    \bottomrule
  \end{tabular}
\end{table}

Despite employing CoG to steer the VLM’s Chain-of-Grounding reasoning as described in Sec. \ref{3.3cog}, we must acknowledge that our system places a heavy inference burden on the VLM, pushing its capabilities to the limit of current state-of-the-art models. To investigate how model choice affects overall system performance, we designed an exploratory experiment using all tasks from the comparative study in Sec. \ref{4.1db}. Evaluation metrics include: (1) the human-evaluated validity of the constraint code and tool selections produced by a single complete CoG inference; and (2) the overall task success rate.

Experimental results are presented in Tab. \ref{tabVLM}. We observe that lower-tier models achieve overall acceptable performance, demonstrating our method’s robustness to model variation, yet more advanced, newer models deliver clearly superior results across both evaluation metrics. This finding conveys two insights. First, our approach will continue to benefit as VLM reasoning capabilities improve in the future. Second, there is a promising research direction in designing system architectures that more effectively structure and guide the VLM’s reasoning process so as to lessen dependence on its peak performance. For these reasons, we ultimately selected GPT-4.1 as the VLM for our method: it delivers nearly the best reasoning performance while costing only about twenty percent of o3.

\subsection{Details of CoG(Chain of Grounding)}
\label{a52cog}

We design Chain of Grounding (CoG) to explicitly guide the VLM’s reasoning : starting from the natural‐language instruction \(I\) and scene observation \(X\), CoG decomposes the task, generates spatial constraints, selects optimal extraction tools, and generate executable constraint functions.  First, CoG performs multi‐stage decomposition:  
\[
\bigl\{\{\mathrm{hints}_{s,j}\}_{j=1}^{J_s}\bigr\}_{s=1}^S = \mathcal{D}(I,X)
\]
where \(S\) is the number of stages produced by the decomposition, and for each stage \(s\), \(\{\mathrm{hints}_{s,j}\}_{j=1}^{J_s}\) is the set of \(J_s\) concise, representation‐agnostic natural‐language operational hints generated to guide spatial reasoning at stage \(s\).  Next, for each stage \(s\) it generates a set of natural‐language spatial constraints:  
\[
C_s
=\;\{\{(c_{s,j,k},O_{s,j,k})\}_{k=1}^{K_{s,j}}\}_{j=1}^{J_s}
=\;\mathcal{C}(\{\mathrm{hints}_{s,j}\}_{j=1}^{J_s})
\]
where \(C_s\) is the set of all spatial constraints for stage \(s\), each constraint \(c_{s,j,k}\) is a natural‐language specification derived from \(\mathrm{hints}_{s,j}\), and \(O_{s,j,k}\) denotes the set of task‐relevant objects involved in constraint \(c_{s,j,k}\); \(\mathcal{C}\) is the operator mapping hints to grounded constraint–object pairs.  Then, for each object \(o\) in stage \(s\), CoG selects the optimal extraction tool from the registry \(\mathcal{R}\):  
\[
T_{s,o} = \mathcal{S}(o,\,C_s,\,\mathcal{R}),\quad o\in O_s
\]
where \(O_s = \bigcup_{j,k}O_{s,j,k}\) is the set of all objects involved in any constraint at stage \(s\), for each \(o\), we prompt the VLM selects the simplest tool \(T_{s,o}\in\mathcal{R}\) whose extracted spatial representation fully satisfies the task requirements for object \(o\). Finally, CoG emits Python‐encoded constraint functions for every constraint \(c\in C_s\):  
\[
F_s = \{\mathcal{E}(c,\{T_{s,o}\}_{o\in O_s})\mid c\in C_s\}
\]
where \(F_s\) is the set of Python functions implementing each constraint \(c\in C_s\) for stage \(s\). For each \(c\in C_s\), \(\mathcal{E}\) generates a Python function that takes as input the spatial representations produced by the selected tools \(\{T_{s,o}\}_{o\in O_s}\) and returns a scalar cost quantifying how well constraint \(c\) is satisfied. In sum, we formalize the entire Chain of Grounding process as
\[
\{(\{T_{s,o}\}_{o\in O_s},\,F_s)\}_{s=1}^S \;=\; \mathscr{G}(I, X),
\]
which grounds the natural‐language instruction \(I\) and scene observation \(X\) into per‐stage tool selections \(T_{s,o}\) for each task‐relevant object and into a series of executable constraint functions \(F_s\) at each stage.

\subsection{Details of ToolKit}

\subsubsection{Samples of the Toolkit we provide}
\label{a61toolkitsamp}

\begin{lstlisting}[language=, caption={Example of Toolkit Registry}, label={lst:mdinput}]
{
  "tool_registry": [
    {
      "tool_name": "CenterPointExtractor",
      "output": "visual_center_points : coordinates of each object's visual center in the camera frame",
      "output_format": [
        {
          "x": "float",
          "y": "float",
          "z": "float"
        },
        "..."
      ],
      "input": ["rgb_image", "depth_image", "object_list"],
      "implementation_summary": "Use Grounding DINO to detect bounding boxes, then compute each box's center point.",
      "avg_historical_exec_time_s": 1.2
    },
    {
      "tool_name": "SemanticKeypointExtractor",
      "output": "object_keypoints : semantically informed keypoints on each object in object_list, in the camera coordinate frame",
      "output_format": [
        {
          "x": "float",
          "y": "float",
          "z": "float"
        },
        "..."
      ],
      "input": ["rgb_image", "depth_image", "object_list"],
      "implementation_summary": "Combine Grounding DINO, DINOv2, SAM, and K-means clustering to extract semantically informed keypoints.",
      "avg_historical_exec_time_s": 3.2
    },
    {
      "tool_name": "UniformSceneKeypoints",
      "output": "uniform_scene_keypoints : points uniformly distributed over objects in the image",
      "output_format": [
        {
          "x": "float",
          "y": "float",
          "z": "float"
        },
        "..."
      ],
      "input": ["rgb_image", "depth_image", "object_list"],
      "implementation_summary": "Use a SAM-based framework (SoM) to sample uniformly distributed keypoints over most objects in the scene.",
      "avg_historical_exec_time_s": 2.3
    },
    {
      "tool_name": "SE3PoseEstimator",
      "output": "6d_poses of object",
      "output_format": { "type": "float", "shape": [4, 4] },
      "input": ["rgb_image", "depth_image", "predefined_object_id"],
      "implementation_summary": "Apply SAM6D to estimate the SE(3) 6D pose for a specified object.",
      "avg_historical_exec_time_s": 5.2
    },
    {
      "tool_name": "LocalSubImageExtractor",
      "output": "a sub_image around the object",
      "output_format": { "type": "RGB_image", "shape": ["H", "W", 3] }
      "input": ["rgb_image", "object"],
      "implementation_summary": "Detect object box with Grounding DINO, expand by padding, and crop the resulting sub-image.",
      "avg_historical_exec_time_s": 1.4
    },
    {
      "tool_name": "VLMTaskPointExtractor",
      "output": "task_relevant_points",
      "output_format": [
        {
          "x": "float",
          "y": "float",
          "z": "float"
        },
        "..."
      ],
      "input": ["rgb_image", "instruction"],
      "implementation_summary": "Prompt VLM to generate extraction code that invokes multiple basic vision tools based on image and instruction.",
      "avg_historical_exec_time_s": 4.1
    },
    {
      "tool_name": "VLMTaskVectorExtractor",
      "output": "task_relevant_vectors",
      "output_format": [
        {
          "start_point": { "x": "float", "y": "float", "z": "float" },
          "end_point":   { "x": "float", "y": "float", "z": "float" }
        },
        "..."
      ]
      "input": ["rgb_image", "instruction"],
      "implementation_summary": "Prompt VLM to generate extraction code for ordered point pairs that represent task-relevant vectors.",
      "avg_historical_exec_time_s": 4.0
    }
    {
      "tool_name": "VLMTaskStateMachineExtractor",
      "output": "task_relevant_state_machine : a finite state machine describing the possible states and transitions of each task relevant object (e.g., {\"drawer\": [\"closed\",\"open\"]})",
      "output_format": {
        "type": "dict",
        "shape": {
          "object_id": ["state1", "state2", "..."],
          "transitions": [
            {
              "from": "state",
              "to": "state",
              "trigger": "string"
            },
            "..."
          ]
        }
      },
      "input": ["rgb_image", "instruction"],
      "implementation_summary": "Prompt the VLM to directly infer and output a state machine model for each task relevant object mapping its discrete states (e.g., open/closed) and allowed transitions based on the current scene image and the user instruction.",
      "avg_historical_exec_time_s": 4.5
    }
    {
      "tool_name": "VLMTaskTopoSorter",
      "output": "task_relevant_topo_order : an ordered list of object IDs sorted by their extraction priority (from most exposed to most occluded)",
      "output_format": {
        "type": "array",
        "shape": ["object_id_1", "object_id_2", "..."]
      },
      "input": ["rgb_image", "instruction"],
      "implementation_summary": "Prompt the VLM to analyze the scene image and instruction, infer each task-relevant object's topological extraction order based on occlusion and stacking relationships, and return a prioritized list of object IDs.",
      "avg_historical_exec_time_s": 3.8
    }
  ]
}
\end{lstlisting}

In this appendix, we briefly introduce the various tools in our Toolkit for spatial representation extraction: CenterPointExtractor computes and outputs each object’s center point coordinates in the camera frame, based on bounding boxes detected by Grounding DINO\cite{liu2024grounding}; SemanticKeypointExtractor combines Grounding DINO, DINOv2, SAM, and K-means clustering to extract semantically meaningful keypoints on each object in the provided list\cite{liu2024grounding,oquab2023dinov2,zou2023segment}, returning their 3D coordinates in the camera frame; UniformSceneKeypoints uses a SAM-based SoM\cite{yang2023set} framework to uniformly sample keypoints over the surfaces of most objects in the scene, outputting their 3D locations; SE3PoseEstimator applies SAM6D\cite{lin2024sam} to estimate a specified object’s full 6D pose, outputting the corresponding SE(3) transformation matrix (4×4); LocalSubImageExtractor first uses Grounding DINO to detect an object’s bounding box, expands it by a fixed padding, and crops that sub-region from the original image for downstream tools; VLMTaskPointExtractor prompts the VLM to generate code invoking multiple basic vision tools, and—given an image and an instruction—outputs a list of task-relevant 3D points; VLMTaskVectorExtractor similarly leverages the VLM to produce extraction code, outputting ordered point-pair vectors that represent object directions or motion paths; VLMTaskTopoSorter invokes the VLM to directly infer and return a topology-based ordering of task-relevant objects, useful for staged or occluded scenarios; and VLMTaskStateMachineExtractor has the VLM infer a state machine for each task-relevant object (e.g., open vs. closed drawer), supporting operations that depend on object state.

We observe an intriguing direction: leveraging Vision–Language Models (VLMs) as fundamental spatial representation extractors. This idea stems from our recognition that certain basic spatial representations—such as keypoints and vectors—are critical for robotic manipulation, yet existing open-source tools often fall short in one or more aspects (e.g., extraction speed, instruction understanding, ease of deployment). To address this, we explore using VLMs with their rich world knowledge and reasoning capabilities to extract these representations. Our approach is to prompt a VLM, given a natural-language instruction, to infer all low-level vision conditions necessary for extracting a given representation—such as hue segmentation, saturation and brightness thresholds, shape and area filtering, solidity detection, edge and corner detection, specular highlight isolation, etc.—and to automatically generate locally runnable code that composes multiple basic vision operators (e.g., via OpenCV) to perform the extraction, ultimately outputting the desired spatial representation. We find that the generated code is predominantly based on OpenCV routines. This reasoning-plus-code-generation strategy not only achieves keypoint and vector extraction quality comparable to or exceeding specialized vision models, but also fully leverages the VLM’s affordance reasoning priors.

In terms of unconventional representations, we introduce two classes of high-level spatial abstractions beyond traditional constraints: first, object topology graphs that capture stacking, dependency, or occlusion relationships among multiple items in the scene; and second, state-machine representations of task-relevant objects, which reflect each object’s state and its transitions across different task steps. For example, in drawer-manipulation tasks we prompt the VLM to generate the drawer’s open/closed state machine to support subsequent, state-dependent decisions; whereas in fragile-stack-handling tasks we extract a topological ordering of the objects to guide careful handling.  

\begin{figure}[ht]
  \centering
  \includegraphics[width=\linewidth]{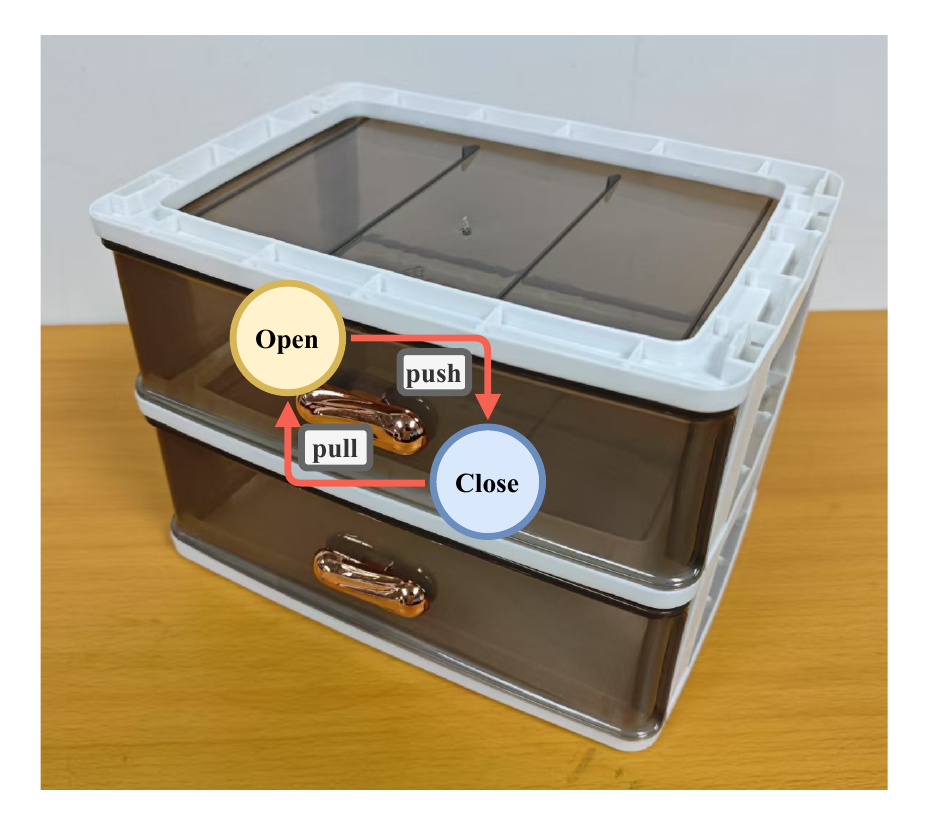}
  \caption{An example of a State Machine. Modeling the drawer’s states to support the generation of state-dependent constraints.}
  \label{fig:diagram}
\end{figure}

\subsubsection{Details and Discussion of 6D Pose Estimation Tools}

6D pose, defined as an object’s position and orientation in three-dimensional space, is a critical spatial representation for robotic manipulation tasks and is indispensable in a wide range of applications. Although numerous 6D pose estimation methods have emerged in recent years, challenges remain in achieving real-time performance, broad generality, and ease of deployment.

Typical inputs to a 6D pose estimation tool are RGB-D data and, optionally, a CAD model of the object; the output is usually an estimate of the object’s pose in SE(3). Methods can be broadly categorized by their dependence on known object models: (1) \textbf{model-based} approaches\cite{lin2024sam,wen2024foundationpose,zhang2024omni6dpose}, which require a pre-provided mesh model (e.g., PLY or STL format); and (2) \textbf{model-free} approaches\cite{liu2022gen6d,sun2022onepose,he2022onepose++}, which may still rely on other forms of prior knowledge—such as an object point cloud—rather than a full mesh.

In our toolkit, we include SAM6D\cite{lin2024sam} as a representative model-based extractor, which requires a 3D model of each target object. To generate these models, we first reconstruct a textured 3D mesh (.glb) from a single RGB image using SPAR3D\cite{huang2025spar3d}, then convert the mesh into the PLY format accepted by SAM6D via lightweight web tools such as ImageToStl. This pipeline provides the high-quality object models needed for accurate 6D pose estimation.

While single-image 3D reconstruction (e.g., with SRAR3D) is a promising research direction, it currently suffers from quality issues such as incomplete backside reconstruction. Exploring more robust real-time reconstruction tools represents a viable avenue for future work. We also acknowledge that the requirement for object priors (e.g., full mesh models or point clouds) limits the system’s ability to handle previously unseen objects. As more reliable, real-time, prior-free 6D pose estimation methods emerge, our modular T-Rex framework will seamlessly integrate them to further enhance performance.  

\begin{figure}[ht]
  \centering
  \includegraphics[width=\linewidth]{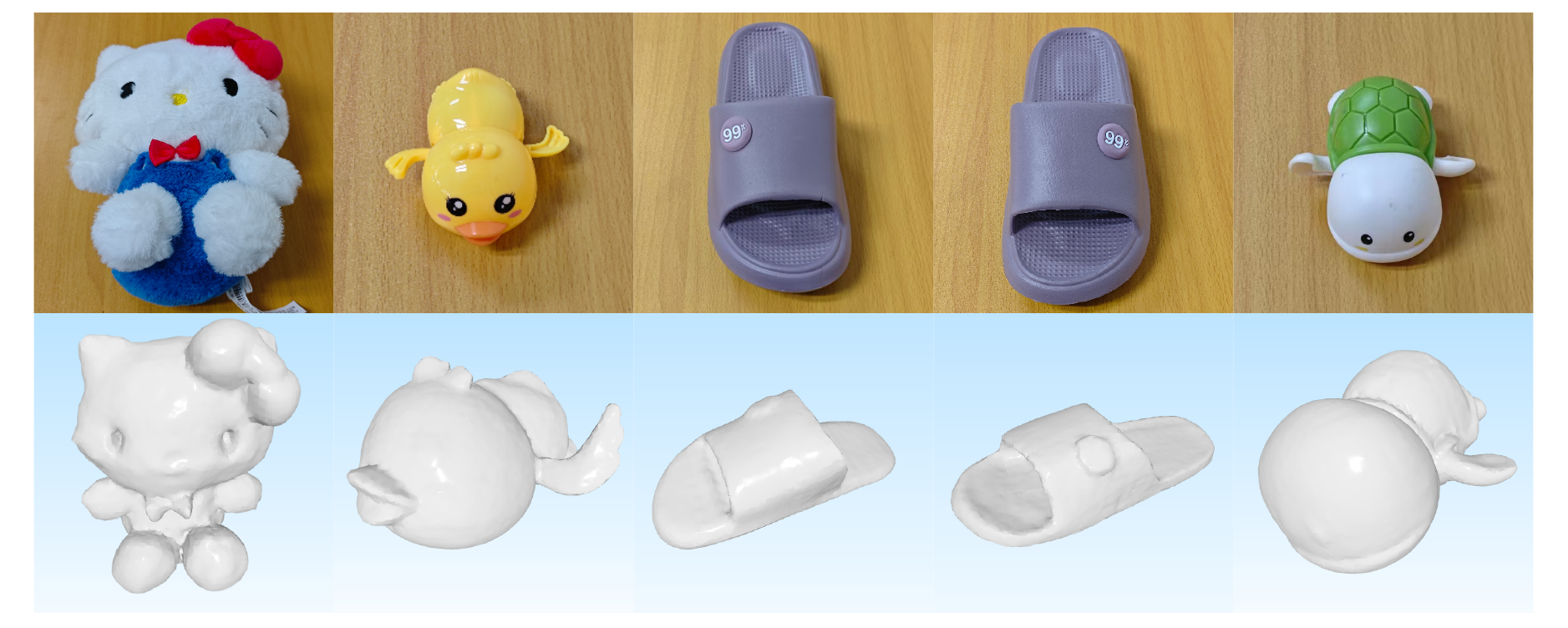}
  \caption{Example experimental props and the 3D meshes we reconstructed}
  \label{fig:diagram}
\end{figure}

\subsubsection{Discussion on other representation that can be integrated into the pipeline}

Some modules from other works can also be seamlessly integrated into our system as spatial-representation extractors\cite{pan2025omnimanip,huang2023voxposer,huang2024copa,liu2024moka}. These modules provide additional perception capabilities to enhance overall performance.

Although the computer-vision community has produced a wealth of techniques for various vision tasks, they often face significant challenges when directly applied to robotic manipulation. On one hand, robots demand vision modules that are real-time, robust, and lightweight, whereas many vision algorithms rely on complex priors or multi-stage pipelines that struggle to meet such requirements. On the other hand, building and maintaining these priors (e.g.,  object 3D meshes) requires substantial engineering effort. We believe that adapting state-of-the-art computer-vision advances for robotic manipulation is a highly valuable research direction, and we encourage readers to explore more efficient and scalable solutions in this space.

Beyond visual information, tactile feedback can directly sense contact forces\cite{lin20239dtact,zhang2022tac3d}, which is critical for fine-grained grasping tasks. Such sensory data itself constitutes a form of representation that can be fed in parallel with visual representation into the T-Rex framework, allowing the construction of constraints and generation of robot action sequences to provide the system with richer perceptual modalities.

\subsubsection{Details of Representation Tracking}
\label{a64track}

For some fundamental computer vision functions, such as object detection\cite{wang2025yoloe,tian2025yolov12,wang2024yolov10}, current methods already support robust video‐streamed detection and tracking with good temporal consistency, making them well suited for robotic manipulation tasks.

However, for more advanced spatial representations (e.g., vectors or 6D poses), existing vision techniques are not yet as mature. Robotic manipulation often requires approximate real‐time, video‐stream-level tracking, ensuring that the interval between successive extractions is below the task’s acceptable tracking latency and that results remain consistent across frames.

Some works\cite{zheng2023pointodyssey,karaev2024cotracker,vecerik2024robotap,xiao2024spatialtracker} investigate using a separate tracking module to follow the outputs of a preceding extraction stage.  Though promising, such approaches can become bottlenecks in practical robotic systems in terms of inference latency, computational overhead, memory footprint, and overall system complexity.

To approximate continuous tracking with our existing tools, we provide a high-frequency invocation interface in our framework.  This allows re-extracting spatial representations on key frames or at fixed intervals—e.g., repeatedly calling `CenterPointExtractor` or `SemanticKeypointExtractor` during execution.  This strategy works well for simple tasks with low tracking demands (e.g., object classification or coarse position correction), but for complex tasks requiring fine-grained, high-frequency tracking, the average execution time per tool call invariably exceeds the ideal frame interval.  Consequently, tracking precision and real-time performance suffer, and spatial representation tracking can become a system bottleneck—consistent with observations in related work.

\subsection{Details of Low‑Level Action Sequence Generation}
\label{low}

Low‐level action sequence generation is the final stage of our pipeline. Its primary role is to convert the spatial representations and constraint functions produced by CoG and the representation extraction modules into concrete action sequences that the robot can execute. This module takes, as input, the Python constraint functions corresponding to each stage’s natural‐language constraints and the associated object spatial representations, and outputs trajectories in SE(3) along with grasp/release commands, which together drive the robot to complete the manipulation task.

In robotic manipulation tasks, representations serve as the crucial bridge between real‐world scenes and the reasoning capabilities of Vision‐Language Models (VLMs), acting as the primary handle by which VLMs perform grounded inference. Although a comprehensive taxonomy of representations in robotic manipulation has yet to emerge, we broadly categorize scene representations into two types: (1) conventional representations: those with group structure, such as keypoints, directional vectors, and 6D poses (SE(3)); and (2) non‐conventional representations: discrete attributes that lack group structure, such as state representations (e.g., whether a drawer is open or closed) and topological orderings (e.g., the stacking order of multiple objects). “Non‐conventional” here simply denotes representations that have received relatively little attention in the field so far, not that they are unimportant; for more complex tasks, these representations can be critical. They both offer additional ways to parse scene information and point toward promising directions for future research. Further discussion of non‐conventional representations can be found in the appendix.

Our low‐level action sequence generation module operates as follows: first, if all constraints in this stage are based on conventional geometric representations (e.g., points, vectors, or 6D poses) and these constraints can be efficiently solved via continuous optimization, the system simply invokes the solver, which uses the full set of constraint functions and tracked representations to produce the action sequence. Only when a stage includes constraints built on non‐conventional representations, such as those requiring object state judgments or relying on topological relations, does the code generation mechanism trigger: the system first generates a complete, executable script for that stage, integrating the previously defined constraint functions and low‐level control interfaces into a coherent execution flow. Within that script, sub‐parts corresponding to constraints based on conventional representations defer to the solver to compute the action sequence. This design offers several advantages: (1) adaptability: most scenarios are handled purely by the solver, with no extra code to write or maintain; and (2) layered structure: the solver focuses on continuous motion refinement, while code generation handles decision‐making and logical control flow.

We acknowledge that this small portion of our pipeline---the solver‐based action sequence generation---draws significant design inspiration from the ReKep\cite{huang2024rekep}, and thus we do not claim it as our contribution. We gratefully recognize ReKep’s outstanding impact on the field of robotic manipulation. For further details on the solver‐based action sequence generation component, we encourage readers to consult ReKep’s publications.

We further acknowledge that this constraint‐adaptive low‐level strategy generation is not necessarily the ultimate solution. In principle, one could devise a more powerful low‐level planner that uniformly handles constraints derived from all representation types as inputs. However, since this lies beyond our core contribution (task‐adaptive representations), and our current approach already performs well enough to demonstrate our key insights, we do not explore it here—though we recognize it as a valuable avenue for future improvement. We encourage readers to investigate such unified low‐level planning methods, which we believe hold great promise.

\subsection{Extended Discussion of Limitations and Future Works}

\label{limi}

While advantageous, T-Rex also has limitations. First, its performance is to some extent constrained by the current state of research on Large Vision Models; as an illustration, it remains challenging to maintain precise spatial representation tracking under conditions of intermittent occlusion. Second, our approach considers only end-effector trajectories, whereas whole-arm planning is clearly a better design choice, as it can more effectively handle collision avoidance and related issues. Third, as a demonstration, our Toolkit currently includes only a limited number of selectively chosen representation extraction tools for spatial representation extraction. Under these conditions, directly providing the Toolkit Registry to the VLM for tool selection is feasible. However, as the Toolkit is scalable and the number of available tools increases significantly, the CoG and context learning-based tool selection approach may face challenges in terms of inference efficiency and accuracy.

The rapid development of Large Vision Models and Vision-Language Models opens several exciting avenues for future work. For example, by collecting and constructing large-scale datasets and fine-tuning existing VLMs, it may be possible to develop Embodied Reasoning Models with enhanced capabilities in spatial reasoning and affordance inference. Although data collection and training processes are resource-intensive, they represent the most direct path toward endowing VLMs with embodied capabilities. In addition, as advancements in computer vision research continue, it is anticipated that increasingly powerful Large Vision Models will emerge, enabling the extraction of a broader range and higher quality spatial representations.

\subsection{Connections and Distinctions between T-Rex and existing works
}
\label{arw}

\paragraph{Vision-Language Models for Robotics.}
The Key Difference between our method and existing approaches is that existing approaches rely on Fixed Types and Granularities of spatial representations, making it difficult to simultaneously meet diverse task requirements and maintain computational efficiency. Our approach introduces Task-Adaptive spatial representation Types and Granularities, and achieves a favorable trade-off between spatial understanding and overall system efficiency.

\paragraph{Grounding Language Instructions.} 
Since VLMs are typically trained on dialogue tasks, there is a gap in their reasoning abilities when applied to robotic tasks. Without additional guidance, their reasoning in embodied tasks sometimes misses or deviates from the correct path. This issue becomes particularly evident in tasks that require long-chain reasoning, especially after our method introduces adaptive spatial representations extends the reasoning chain of VLMs. To address this, we propose CoG (Chain of Grounding) to guide VLMs step by step in mapping instructions, ultimately obtaining comprehensive and correct spatial constraints for task planning at each stage, and selecting the most appropriate spatial representations for each entity at each stage to complete the task.

\paragraph{Vision Foundation Models and Spatial Representation for Robotics.}   
Compared to existing methods that rely on a fixed extraction pipeline to produce a single type of spatial representation, our approach adaptively selects the most appropriate representation types and extraction methods based on specific task requirements, effectively balancing representational power and runtime efficiency.

\subsection{Further formal discussion}
\label{formal}

Our framework adopts the classical pipeline of representation extraction, constraint generation and low‑level solving. First, under the guidance of the Chain of Grounding, the VLM infers the spatial representation extraction tools seletions \(\{T_{s,o}^*\}\) and the set of executable Python constraint functions \(\{F_s\}\) for each stage \(s\), formalized as
\[
\{(\{T_{s,o}^*\}_{o\in O_s},\,F_s)\}_{s=1}^S \;=\;\mathscr{G}(\ell,X_0).
\]
Next, using the selected tools \(\{T_{s,o}^*\}\), the Spatial Representation Extraction Toolkit \(\mathcal{R}\) extracts per‑stage, per‑object representations \(\{r_{s,o}\}\) via
\[
r_{s,o} \;=\; T_{s,o}^*(\ell,X_0).
\]
Finally, the low‑level action sequence generator \(\Pi\) consumes the constraint functions \(\{F_s\}\) and the extracted representations \(\{r_{s,o}\}\) to produce the  robot action sequence:
\[
\{\tau_i\}_{i=1}^N \;=\;\Pi\bigl(\{F_s\},\{r_{s,o}\}\bigr).
\]
In this way, the resulting trajectory set \(\{\tau_i\}\) transforms the scene \(X\) into one that satisfies the instruction \(\ell\).

The system’s core components such as the Spatial Representation Extraction Toolkit and the Chain of Grounding are presented comprehensively in Sec. \ref{3.2ta} and \ref{3.3cog}. Other components of the system are detailed in the appendix such as the Low-Level Action Sequence Generator(see \ref{low}) and Spatial Representation Tracking (see \ref{a64track}).

\subsection{Broader Impacts}
\label{Broader}

T-Rex can be used in manufacturing to improve the accuracy and flexibility of automated production and reduce repetitive labor. T-Rex enhances robot’s spatial reasoning capabilities, but it could be repurposed for negative applications such as advanced autonomous weapon development or pose safety risks if VLM inferences are flawed. To mitigate these hazards, we recommend incorporating human-in-the-loop verification for high-risk tasks and establishing deployment monitoring and feedback mechanisms to ensure responsible use.


\newpage
\end{document}